\pdfoutput=1
\documentclass[11pt]{article}

\usepackage[final]{acl} 
\usepackage{times}
\usepackage{latexsym}

\usepackage[T1]{fontenc}
\usepackage[utf8]{inputenc}

\usepackage{microtype}

\usepackage{inconsolata}

\usepackage{graphicx}
\usepackage{algorithm}
\usepackage{algpseudocode}
\usepackage{wrapfig}
\usepackage{booktabs}  % f
\usepackage{subcaption}
\usepackage{tikz}
\usetikzlibrary{positioning, arrows.meta}

\usepackage{amsmath,amsfonts,bm}

\usepackage{amsthm}
\theoremstyle{plain}

\theoremstyle{definition}

\theoremstyle{remark}

\def\figref#1{figure~\ref{#1}}
\def\Figref#1{Figure~\ref{#1}}

\def\eqref#1{equation~\ref{#1}}
\def\1{\bm{1}}

\def\ve{{\bm{e}}}

\def\vh{{\bm{h}}}

\def\vx{{\bm{x}}}

\DeclareMathAlphabet{\mathsfit}{\encodingdefault}{\sfdefault}{m}{sl}
\SetMathAlphabet{\mathsfit}{bold}{\encodingdefault}{\sfdefault}{bx}{n}

\usepackage{scalerel,stackengine}
\stackMath
\newcommand\reallywidehat[1]{%
\savestack{\tmpbox}{\stretchto{%
  \scaleto{%
    \scalerel*[\widthof{\ensuremath{#1}}]{\kern-.6pt\bigwedge\kern-.6pt}%
    {\rule[-\textheight/2]{1ex}{\textheight}}%WIDTH-LIMITED BIG WEDGE
  }{\textheight}% 
}{0.5ex}}%
\stackon[1pt]{#1}{\tmpbox}%
}

\title{From Characters to Tokens: Dynamic Grouping with Hierarchical BPE}

\author{
  \textbf{Rares Dolga\textsuperscript{1,2}},
  \textbf{Lucas Maystre\textsuperscript{2}},
  \textbf{Tudor Berariu\textsuperscript{2}},
  \textbf{David Barber\textsuperscript{1,2}}
\\
\\
  \textsuperscript{1}University College London, UK \\
  \textsuperscript{2}UiPath, UK
\\
  \small{
    \textbf{Correspondence:} \href{mailto:rares.dolga@ucl.ac.uk}{rares.dolga@uipath.com} 
  }
}

\begin{document}
\maketitle
\begin{abstract}
Subword tokenization methods like Byte Pair Encoding (BPE) are widely used in large language models due to their balance of vocabulary compactness and representational power. However, they suffer from inefficiencies in representing rare words and require large embedding matrices. Character-level models address these issues but introduce performance bottlenecks, particularly in Transformer-based architectures. Recent hierarchical models attempt to merge the benefits of both paradigms by grouping characters into patches, but existing patching strategies either rely on whitespace—limiting applicability to certain languages—or require auxiliary models that introduce new dependencies. In this paper, we propose a dynamic character grouping method that leverages the structure of existing BPE tokenization without requiring additional models. By appending explicit end-of-patch markers to BPE tokens and introducing a second-level BPE compression stage to control patch granularity, our method offers efficient, flexible, and language-agnostic representations. Empirical results demonstrate that our approach matches or exceeds the performance of dynamic entropy- and whitespace-based patching strategies, while maintaining a compact vocabulary.
\end{abstract}

\section{Introduction}
% \begin{itemize}
%     \item Our secondary bpe can be applied on top of space tokeizers as well + footnote test
%     \item Same for quick decoding.
% \end{itemize}

% \begin{itemize}
%     \item Say you have a learning stage as well - but now weights (bpe algo)
%     \item Empshasze more that people did not try this patching because of eos per patch which increases length. But we get around it.
%     \item Say space is good if we ensure same patch - Experiment with no max len. 
% \end{itemize}
Subword tokenization algorithms, particularly Byte Pair Encoding (BPE), have become the de facto standard for text representation in large language models due to their balance between vocabulary compactness and representational flexibility. However, despite their widespread use, subword methods introduce notable limitations. Embedding matrices tied to large vocabularies become parameter inefficient, for which rare words often have bad representations. While BPE provides a degree of compression over raw byte sequences—enhancing computational efficiency, the achievable compression is fundamentally constrained by the vocabulary size. For example, modern tokenizers used in models such as Gemma 2 and LLaMA 3 employ vocabularies of around 250K tokens, which inherently limits the extent to which sequence length can be reduced.

\begin{figure}
    \centering
    \includegraphics[width=\linewidth]{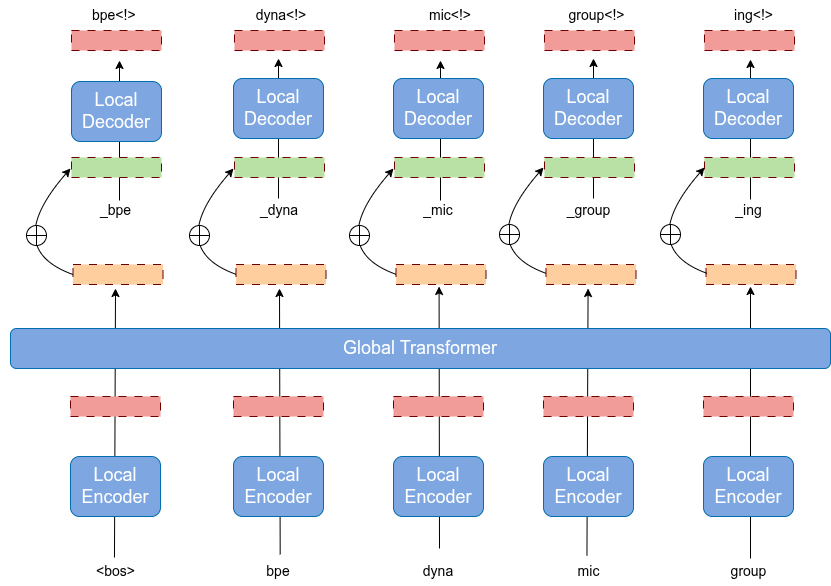}
    \caption{Hierarchical Model: Next patch prediction with autoregressive character prediction per patch. A patch is given by a BPE token.``<!>'' represent end of token.}
    \label{fig:hirachical-model}
\end{figure}

Character-level models directly address many of the limitations inherent in subword-based tokenization, particularly with regard to rare word handling and parameter efficiency. However, this comes at the cost of runtime performance, especially in quadratic-complexity architectures such as standard Transformers. To bridge this gap, recent work has explored hierarchical representations that aim to combine the flexibility of character-level input with the efficiency of subword models. These approaches typically group characters into larger units—referred to as ``patches''—and learn representations via neural networks rather than embedding tables.

The effectiveness of such models is closely tied to the patching strategy. Naively grouping consecutive characters has been shown to underperform relative to traditional BPE tokenization \cite{metabyte, spacebyte}. In contrast, dynamic grouping strategies, particularly those that treat whitespace as a delimiter, have achieved competitive or state-of-the-art results. However, whitespace-based segmentation does not generalize to logographic writing systems such as Chinese, where spacing is not semantically meaningful.

One proposed solution involves training a lightweight character-level model to identify patch boundaries based on information-theoretic criteria, such as local entropy. In this setup, new patches are initiated in regions of high entropy (i.e., high surprise). While promising, this approach introduces a dependency on an additional model, which may be sensitive to domain shifts and data variability.

In this paper, we address the limitations of character-level and subword-based models by introducing a dynamic character grouping method that avoids the need for training an additional model to determine patch boundaries. Our approach defines each patch as the sequence of characters that constitute a BPE token, effectively repurposing the tokenization process itself as a grouping mechanism. To enable incremental processing, we modify the standard BPE encoder by appending an explicit end-of-patch marker after each token. While prior work has briefly mentioned this idea, to the best of our knowledge, we are the first to provide a detailed empirical evaluation.

% To address the fact that we increase the cost by appending an end of patch symbol to each patch, we provide a hierarchical BPE algorithm to compress each patch of characters to a maximum predefined size $S$.  This hierarchical compression exploits the observation that small n-grams are frequently repeated across tokens, enabling effective merging and reduced patch lengths. Our method can thus be interpreted as a form of \emph{dynamic grouping of tokens}, rather than at character level.

To offset the additional cost introduced by appending an end-of-patch symbol to each patch, we introduce a hierarchical BPE algorithm that compresses each character-level patch to a maximum predefined length $S$. This compression leverages the observation that short $n$-grams appear frequently across tokens, allowing for effective merging and shorter patch lengths. Consequently, our method can be viewed as performing \emph{dynamic grouping of tokens}, rather than operating solely at the character level.

Empirically, our model outperforms entropy-based patching and achieves comparable performance with whitespace-based dynamic grouping, while maintaining broader applicability across writing systems. Moreover, compared to standard BPE where the vocabulary is a large, our approach yields better efficiency, requiring significantly fewer FLOPs.

\section{Methodology}
% We use a hierarchical model which allows us to better exploit the tradeoff between several parameters and runtime efficiency. The input consists of a sequence of characters $c_1, c_3, \dots c_L$, which we tokenise with a pre-trained BPE tokeniser to obtain a sequence of tokens $\vx_1^{1:S}, \dots ,\vx_T^{1:S}$. Each token becomes a dynamic patch, with a maximum of $S$ characters\footnote{We pad to $S$ if the patch has fewer characters}, i.e $\vx_t^{1:S} = (c_t^1, c_t^2, \dots c_t^S)$.

We introduce a hierarchical representation model that enables more effective trade-offs between granularity, sequence length, and run-time efficiency. Given an input sequence of characters \( c_1, c_2, \dots, c_L \), we apply a pre-trained BPE tokeniser to produce a sequence of variable-length subword tokens $\vx_1, \dots ,\vx_T$. Each token is sequence of characters, which we pass to our hierarchical BPE algorithm. Our algorithm takes the sequence of characters and compresses it to a shorter sequence of integers and adds an end of sequence marker. Then the sequence representing the initial BPE token is padded to the maximum length $S$\footnote{One can reduce the amount of padding by concatenating the patches and applying a sliding window approach. A mask can be used to delimit the sequences.} and passed to the neural network detailed in section~\ref{sec:herarch-model}.
% Each token is then assigned to a \emph{dynamic patch}, represented as a character sequence \( \vx_t^{1:S} = (c_t^1, c_t^2, \dots, c_t^S) \), where \( S \) is a predefined maximum patch length. Tokens shorter than \( S \) characters are zero-padded to maintain consistent dimensionality across the sequence. 

\begin{algorithm}[t]
\caption{Hierarchical BPE}
\label{alg:hierarchical-bpe}
\begin{algorithmic}[1]
\Require Maximum patch size $S$
\Require Tokens $\mathcal{T} = \{ \vx_v^{1:S} \mid v \in \{1, \dots, V\} \}$
\State $\texttt{merges} \gets [\,]$
\State $V'\gets 0$
\While{$|\mathcal{T}| > 0$}
    \State $\texttt{pair} \gets \textsc{MostFreqPair}(\mathcal{T})$
    \State $V' \gets V' + 1$
    \State $\mathcal{P} \gets \textsc{Merge}(\texttt{pair}, \texttt{merges}, \mathcal{T})$
    \ForAll{$t \in \mathcal{P}$}
        \If{$\texttt{len}(t) \leq S$}
            \State Remove $t$ from $\mathcal{T}$
        \EndIf
    \EndFor
\EndWhile
\State \Return $V', \texttt{merges}$
\end{algorithmic}
\end{algorithm}

\subsection{Hierarchical BPE}
\begin{figure}[!h]
    \centering
    \includegraphics[width=0.4\textwidth]{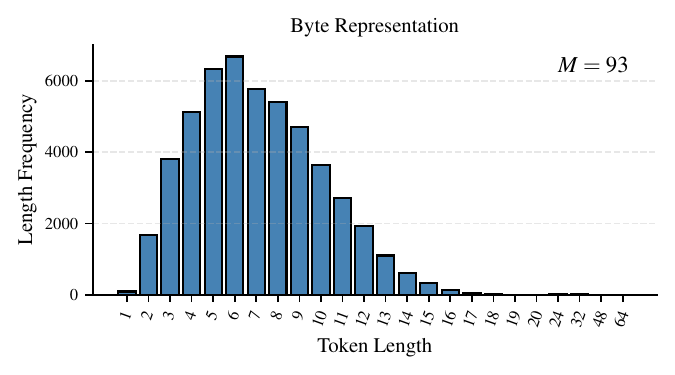}
    \caption{Histogram of lengths for all tokens in the GPT2 tokenizer. The first plot is truncated on the x-axis, from maximum token length $M=93$. }
    \label{fig:histo-freq}
\end{figure}

While representing BPE tokens as sequences of characters which are modelled by local networks effectively leverages syntactic correlations and handles rare tokens, we observe notable computational inefficiencies. 
An analysis of the GPT-2 tokenizer ( a pre-trained BPE tokenizer, reveals that although the longest token spans 93 bytes, the distribution of token lengths is highly skewed, with the majority containing fewer than 15 bytes (\figref{fig:histo-freq}, left). This skew leads to memory spikes in the local models. Additionally, appending a delimiter (e.g., “<>”) to each patch increases patch length, further slowing down the decoding process.

To address these issues, we propose a novel hierarchical tokenization strategy that incorporates a secondary BPE algorithm. This second-stage tokenizer operates on character sequences derived from the initial BPE tokens. Specifically, the algorithm identifies all tokens shorter than a predefined threshold $S$ (the maximum patch length), selects the most frequent byte pair among them, and merges the pair into a new symbol. This procedure is repeated until all tokens conform to the length constraint $S$. The full algorithm in described in \figref{alg:hierarchical-bpe}.

The algorithm can be easily understood with an example. Let us assume a pre-trained BPE tokenizer splits the text ``This is a test!'' into four tokens as in \figref{fig:hirachical-model}. Also, we set $S=6$. The algorithm then looks at the entire vocabulary and sees that ``is'' the most frequent pair\footnote{If other pairs are more frequent, they are used first.}, therefore merging the pair into a new symbol with encoding ``257''. The representation of ``tok1'' becomes $(84, 104, 257, 32, 257, 256)$, where 84, 104 and 32 are the ASCII representations of ``T'', ``h'' and space. 256 represents the end of patch marker. Since the length of the representation for ``tok1'' is smaller than or equal to $S$, this token will not be further compressed.

\begin{figure}
    \centering
    \includegraphics[width=0.8\linewidth]{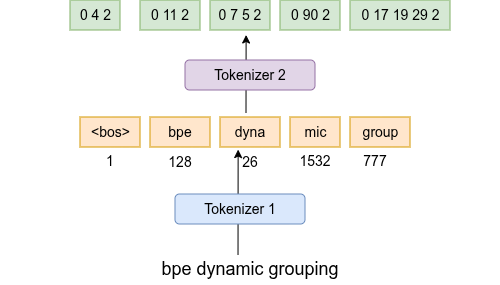}
    \caption{Hierarchical Tokenisation at test time.}
    \label{fig:hierarchical-bpe}
\end{figure}

Our hierarchical BPE framework can also be viewed as a mechanism for \emph{dynamic grouping} over tokens from the vocabulary $V'$. Concretely, a sequence of tokens $x_t \in V'$ can be grouped into patches using a secondary, pre-trained BPE tokenizer with a much larger vocabulary $V \gg V'$. This approach enables the model to dynamically form higher-order token groupings, capturing richer structure in the data while maintaining computational efficiency. An intuitive view of how the hierarchical tokenisation works at test time is presented in \figref{fig:hierarchical-bpe}.

\subsection{Hierarchical Model \label{sec:herarch-model}}
We give an overview of the hierarchical model in \figref{fig:hirachical-model}. More precisely, the model can be decomposed into a local encoder $g_{\phi}$, a latent transformer $f_{\theta}$ and a local decoder $m_{\psi}$.
The encoder $f_{\phi}$ independently maps all the sub-sequences to a fixed continuous representation:
\begin{align}
    \ve_t =  g_{\phi}(\vx_t^{1:S}) \in \mathbb{R}^{D}
\end{align}
We subsequently apply a causal \emph{latent transformer} $f_{\theta}$ to the sequence $\ve_1, \dots, \ve_T$, producing the hidden representation:
\begin{align}
\vh_t = f_{\theta}(\ve_{<t}) \in \mathbb{R}^{D'}
\end{align}
Compared to the local encoder and decoder, the latent transformer is substantially larger, dedicating greater computational capacity to modeling the more complex global structure across the sequence of patches.
Finally, we pass the hidden representations to a local autoregressive decoder $m_{\psi}$, which will predict the sub-sequence representation of the next token: 
\begin{align}
    p(x_t^s = c | x_{t}^{<s}, x_{< t}^{1:S}) = \frac{e^{m_{\psi}(\vx_{t}^{<s}, \vh_{t-1})[c]}}{\sum_{v \in \mathbb{V}} e^{m_{\psi}(\vx_{t}^{<s}, \vh_{t-1})[v]}}
\end{align}

% \david{However, our approach introduces a key distinction: instead of using entropy-grouped or consecutive byte patches, we represent each patch as a sequence of characters derived from BPE tokens. This design leverages the linguistic structure captured by BPE.}{I'm not sure it's good to discuss how this relates to other approaches here. I would prefer that you make a really clear description of the method. You can then discuss how it relates to other approaches in a later section.}

% \david{Our training objective simply becomes the autoregressive prediction of characters in the next patch:}{You write the objective as a function of $\mu$ -- this means that you don't learn $\theta$?}
% \begin{equation}
%     L(\mu) = -\sum_{t=1}^T \sum_{s=1}^S \log p_{\mu}(x_t^s | x_{<t}^{1:S}, x_{t}^{<s})
%     \label{eq:train-obj}
% \end{equation}

\subsection{Metrics}
As our experiments involve both subword and character-level sequence segmentations, it is necessary to normalize evaluation metrics to enable meaningful comparison. To this end, we convert token-level perplexities to bits per byte (BPB), based on the standard definition of information content.
Assuming a learned model $p_{\theta}$ and a sequence $x$ of size $|bytes|$, BPB is given by:
\begin{align}
    bpb &= \frac{-\log_2 p_{\theta}(x)}{|bytes|}
\end{align}
One can show the above by considering that the total information content $I(x)$ of the test set remains invariant across segmentation schemes:
\begin{align}
    I(x) =  |toks| \times & \frac{bits}{token} = |bytes| \times \frac{bits}{byte} \\
    \Rightarrow  \frac{bits}{token} &= \frac{|bytes|}{|toks|}  \times bpb \\
    \Rightarrow I(x) =  |toks| & \frac{|bytes|}{|toks|} \times bpb \Rightarrow bpb = \frac{I(x)}{|bytes|}
\end{align}
We can then use the fact that for a data distribution $p_D$ and a model $p_{\theta}$, the information contained in a sample $x$ is:
\begin{align}
    I(x) &= H[p_d, p_{\theta}] = -\log_2 p_{\theta}(x) \\
    \Rightarrow  bpb & = \frac{-\log_2 P_{\theta}(x)}{|bytes|}
\end{align}
In the above, $H$ refers to entropy, $bpb =  \frac{bits}{byte}$, $|toks|$ means the number of tokens in the test set, and $|bytes|$ refers to the number of bytes in the test set.

Finally, we define \textbf{token fertility} as $F = \frac{|\text{tokens}|}{|\text{words}|}$,
which quantifies the degree of compression applied to the \emph{global sequence} on which the \emph{latent model} $f_{\theta}$ operates. When fertility is less than 1, the latent model processes shorter sequences than the original word-level input, leading to increased computational efficiency.

\section{Experiments}

\subsection{Experimental Setup}
% Our experiments focus on language modelling and downstream tasks like question answering from the~\citet{eval-harness} benchmark.
% For language mdelling we use the first two chunks of the Slimpajama datset~\cite{slimpajama}, which we noticed that contains Chinese characters along plain english text. From the model perspective, we focus on 2 sizes: a small 128M base model and a larger one of 350M which we train from 2.5B tokens to 15B tokens. While the exact size varies, depending on the baseline, we choose the size of the Latent Transformer to equal across experiments. We only vary the embeddings which we replace with local models where it is the case. Table~\ref{tab:hyperparams}.

Our experiments target both language modeling and downstream tasks such as question answering, using the evaluation benchmark introduced by~\citet{eval-harness}. For language modeling, we use the first two chunks of the SlimPajama dataset~\cite{slimpajama}, which notably contain Chinese characters interspersed with English text. We evaluate two model sizes: a smaller 123M-parameter baseline and a larger 359M-parameter model, each trained on token counts ranging from 2.5B to 15B. While model sizes vary, the Latent Transformer architecture remains fixed across all experiments. The only architectural modification involves the embedding layers, which are substituted with local models when applicable.
%For simplicity, we omit the local cross-attention layers and use simple transformer layers in all experiments.
A summary of hyperparameter configurations is provided in Table~\ref{tab:hyperparams}. We also release our code with all the experimental configuration\footnote{Anonymised git url.}. Our experiments were run on 4 A100, and take between 4 hours to 24 hours to run, depending on the experiment.

\begin{table}[h]
\centering
\caption{Model hyperparameters for 359M and 123M parameter LLMs}
\begin{tabular}{lcc}
\toprule
\textbf{Hyperparameter} & \textbf{Medium} & \textbf{Small} \\
\midrule
Latent Layers             & 24   & 12   \\
Latent Hidden Dim.        & 1024 & 768  \\
Latent FFN                & 2816 & 2048  \\
Latent Heads              & 16   & 12   \\
Local Hidden Dim         & 512 & 512  \\
Local FFN                & 512 & 512  \\
Local Heads              & 8   & 8   \\
Enc/Dec. Layers           & 3    & 3    \\
Learning Rate            & 6e-4 & 4e-4 \\
\bottomrule
\end{tabular}
\label{tab:hyperparams}
\end{table}

\subsection{Baseline Comparison}
\label{sec:flops}

We begin by comparing our method against four alternative text encoding approaches. The first baseline employs standard Byte Pair Encoding (BPE), implemented using the GPT-2 tokenizer. The second is a character-level model trained on sequences of length 8192. We further evaluate two dynamic patching methods: space-based and entropy-based grouping. For both, we adopt the codebase and experimental setup from~\citet{metabyte}, including the same pre-trained entropy model. Notably, these patching baselines incorporate additional parameters due to the use of hash embeddings in their local encoders—components that our model does not include. However, we do not introduce any cross-attention layers, ensuring that the local encoder architecture remains consistent across all experiments.

\begin{figure}[!h]
    \centering
    \includegraphics[width=0.7\linewidth]{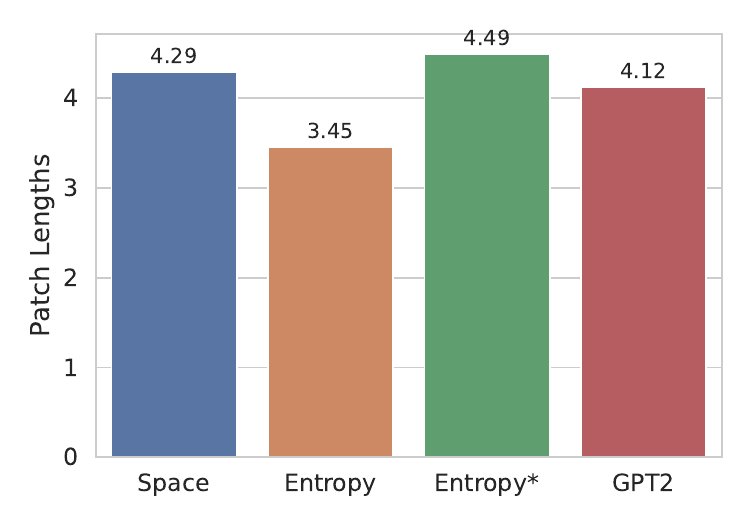}
    \caption{Average Patch Lengths for SlimPajama dataset. "*" represents the unbounded entropy model. }
    \label{fig:patch-lens}
\end{figure}

 In the space-based setting, we enforce a maximum patch size of 6. For the entropy-based method, we explore two configurations: one with the same maximum patch size of 6, and another with no patch size constraint, allowing the model to determine boundaries solely based on local entropy. Interestingly, although the maximum patch size is set to 6, the \emph{actual average patch length} on the SlimPajama dataset is only \emph{4.29} for the space-based method. This can be attributed to the prevalence of short words; by enforcing an upper bound, the average is naturally skewed toward shorter sequences. 
 
 For the entropy-based method, the average patch length is \emph{3.45} in the bounded setting and \emph{4.49} in the unbounded setting. These relatively short patch lengths may be explained by a domain mismatch, as the entropy model was not trained on the SlimPajama dataset. We consider this comparison fair, as our method, based on dynamic grouping using the GPT-2 tokenizer, is also not specifically adapted to SlimPajama. For reference, the GPT-2 tokenizer produces an average patch length of \emph{4.12} characters per token on this dataset. After introducing the end-of-patch marker and applying our second stage BPE with a maximum patch size of 10, the average patch length becomes \emph{4.13}. Figure\ref{fig:patch-lens} summarises the average patch lengths for all the grouping strategies. The average patch length has a direct effect on the number of FLOPs, since it affects the length seen by the big latent model $f_{\theta}$. 

\textbf{FLOPs Calculation.} We estimate the total number of FLOPs in a forward pass based on the average patch length, the local encoder/decoder models, and the latent global model:
\begin{align}
    F &= T \cdot \text{Tr}(p, D_{\text{enc}}, L_{\text{enc}}, V=0) \\
      &+ T \cdot \text{Tr}(p, D_{\text{dec}}, L_{\text{dec}}, V=V') \\
      &+ \text{Tr}(T, D_{G}, L_{G}, V=0)
\end{align}
In the above, $p$ denotes the average patch length, and $T$ is the number of latent tokens, i.e., $T = Y / p$, where $Y$ is the total input length in bytes. $D_x$ represents the hidden dimension, $L_x$ the number of layers, and $V$ the vocabulary size. The function $\text{Tr}(\cdot)$ refers to the standard FLOPs computation for a Transformer model as defined in~\citet{flops-calc}, with the note that embedding operations are assumed to have zero FLOPs.

\begin{table}
    \centering
    % \begin{tabular}{l|cccc}
    \begin{tabular}{l@{\hskip 2pt}|c@{\hskip 3pt}c@{\hskip 3pt}c@{\hskip 3pt}c@{\hskip 3pt}}
    Model &  FLOPs$\downarrow$ & Fertility$\downarrow$ & Params & BPB$\downarrow$ \\
    \toprule
    Ent.-Patch* & 509 & 1.41 & 351M & 1.24 \\
    Ent.-Patch & 668 & 1.83 & 351M & 1.20 \\
    Space-Patch & 534 & 1.45\footnotemark & 351M & 1.14\\
    Char - Level & 4214 & 4.5 & 85M & 1.16 \\
    BPE &  562 & 1.51 & 123M & 1.16 \\
    BPE-Patch & 554 & 1.51 & 99.7M & \textbf{1.11}\\
    % \hline
    % BPE & 15B & 359M & 1.02 & \\
    % Space-Patch & 15B & 575M & 1.04 \\
    % Entropy-Patch & 15B & 575M & 1.10 \\
    % BPE-Patch & 15B & 323M & \textbf{0.98} \\
    % \hline
    \end{tabular}
    \caption{Comparison between our model for S = 10, BPE tokenisation, Space and Entropy patching (Ent.-Patch, Ent.-Space) for 128M models trained on 13500 steps. FLOPs are provided in billions. "*" represents the unbounded model. We report mean over multiple runs on the test dataset. Maximum standard deviation across experiments: $\pm 0.007$.}
    \label{tab:result-1}
\end{table}
%\footnotetext{We split only by space, while the space-patching regex uses additional rules, resulting in a fertility higher than 1.}
\footnotetext{We enforce the maximum space size to be 6, resulting in a fertility higher than 1.}
%\setlength{\tabcolsep}{3pt} % Default value: 6pt
%\renewcommand{\arraystretch}{1.5} % Default value: 1
% \textbf{Results}. Table~\ref{tab:result-1} demonstrates that our approach outperforms all baselines in this setup. Notably, entropy-patching yields the weakest results, likely due to domain mismatch between the entropy model's training data and our dataset. The unbounded model is more efficient, however, the local models need to learn longer patches with fewer parameters, resulting in a big performance drop. When we bound the patch, to a maximum length of 6, the performance increases, at the cost of FLOPs, but it is still far from our model. The space patching model is the most efficient one, and has a good performance. Our model, performs better at the cost of more flops, but the additional advantage of our model is that it can be applied to languages which  do not use space as a word separator. 

% The character model uses less parameters, however is much less efficient at runtime, while the performance lags behind our model. Interestingly, our approach also surpasses standard BPE, despite using fewer parameters. This suggests that the local models in our architecture learn richer representations than those produced by a traditional embedding matrix.

\textbf{Results.} Table~\ref{tab:result-1} shows that our approach outperforms all baselines in this experimental setup. Notably, the entropy-patching method yields the weakest results, likely due to a domain mismatch between the entropy model’s training data and the SlimPajama dataset. While the unbounded entropy model is more efficient, the longer patch lengths require the local models to learn more complex representations with limited capacity, leading to a significant drop in performance. Imposing a maximum patch length of 6 improves results, albeit at a higher FLOPs cost, but the performance still lags behind our method.

The space-based patching model is the most efficient among the baselines and performs reasonably well. Our model achieves the best overall performance, with a higher computational cost. However, it offers a crucial advantage: unlike space-based methods, it generalizes to languages that do not use whitespace as a word separator.

The character-level model uses fewer parameters but suffers from poor runtime efficiency and lower predictive performance. Interestingly, our approach also surpasses standard BPE, despite using fewer parameters. This suggests that the local encoders in our architecture are capable of learning richer token-level representations than those provided by a fixed embedding matrix.

\subsection{Increasing the training time.}
\begin{figure}
    \centering
    \includegraphics[width=0.95\linewidth]{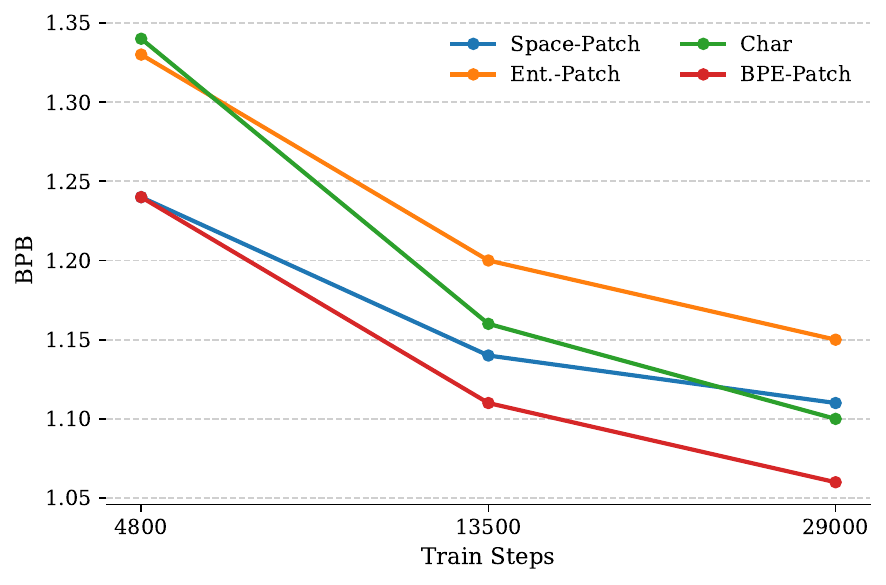}
    \caption{Test performance for different amounts of training.}
    \label{fig:test-len}
\end{figure}

We also examine whether the performance gains of our model persist throughout training. To this end, we compare our approach against character-level modeling as well as space- and entropy-based patching, across different numbers of training steps. We use the same model configurations as in Table~\ref{tab:result-1}. As illustrated in Figure~\ref{fig:test-len}, the BPE-patching method consistently outperforms all baselines as training progresses. Notably, the structure-level representation appears to improve over time, indicating that, in some cases, employing a more granular model—despite its higher computational cost—may lead to better long-term performance.

\subsection{Evaluation on Chinese language}
We evaluate our method on the Skypile dataset~\cite{skypile}, a large-scale Chinese corpus. Small models are trained on a next-byte prediction task for 4,800 steps, and the results are presented in Table~\ref{tab:chinese}. While space-based splitting yields a relatively high patch size, this is due to the use of additional heuristics beyond simple space separation. In contrast, the low patch size observed with the entropy-based tokenizer can be attributed to its neural grouping model, which was not trained on this corpus. Our method achieves the largest patch size—resulting in the lowest FLOPs—while also demonstrating strong performance on this dataset. 

\begin{table}[!h]
\centering
\begin{tabular}{@{}lcc@{}}
\toprule
\textbf{Model} & \textbf{Patch Size}$\uparrow$ & \textbf{BPB}$\downarrow$ \\
\midrule
Space-Patch & 3.06 & 1.24 \\
Entropy-Patch & 1.66 & 1.27 \\
BPE-Patch (LLaMA3) & 3.62 & \textbf{1.20} \\
\bottomrule
\end{tabular}
% \caption{Comparison of different patching strategies on the Skypile data. Llama3 stands for the Llama3 tokenizer used in the fist BPE stage. Patch size refers to average number of bytes per patch in each grouping method.}
\caption{Comparison of different patching strategies on the Skypile dataset. \textit{LLaMA3} refers to the LLaMA3 tokenizer used in the first-stage BPE. \textit{Patch size} indicates the average number of bytes per patch under each grouping method.}
\label{tab:chinese}
\end{table}

\subsection{Increasing the vocabulary}
As previously discussed, increasing the vocabulary size \( V \) improves compression of the sequence input to the global latent model \( f_{\theta} \). Shorter sequences lead to faster runtime, benefiting both the quadratic attention and MLP layers. However, this introduces a tradeoff: as \( V \) increases, the local models must encode more information using fewer tokens, which can strain their capacity. To examine this effect, we conduct an ablation study on vocabulary size. Specifically, we train three SentencePiece tokenizers with vocabulary sizes of 50K, 200K, and 500K, and compare the performance of the standard BPE approach with our model in each setting.

\begin{figure}
    \centering
    \begin{minipage}[t]{0.45\textwidth}
        \centering
        \includegraphics[width=\linewidth]{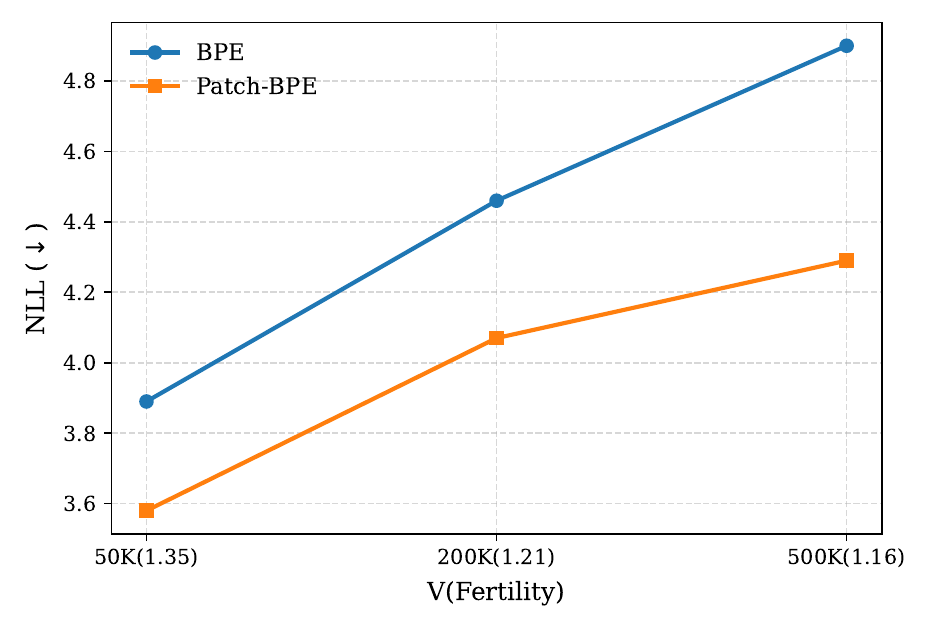}
        \caption{NLL on the test data for varying vocabulary sizes.}
        \label{fig:V-increase-fertility}
    \end{minipage}
    \hfill
    \begin{minipage}[t]{0.45\textwidth}
        \centering
        \includegraphics[width=\linewidth]{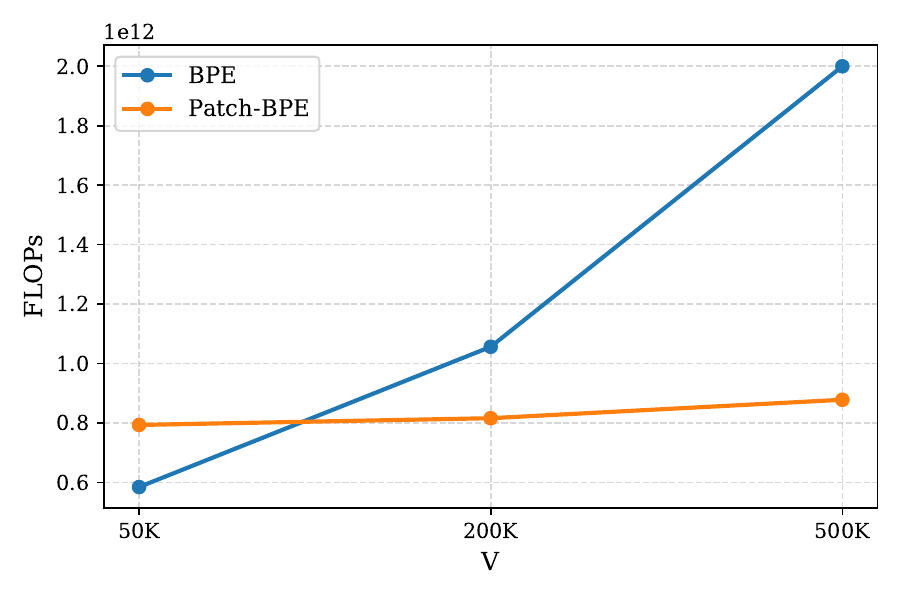}
        \caption{Flops of a forward pass.}
        % \label{fig:plot2}
        \label{fig:flops-increase}
    \end{minipage}
    %\caption{}
\end{figure}

Figure~\ref{fig:V-increase-fertility} shows that the overall performance, measured by the Negative Log Likelihood (NLL), decreases as vocabulary size increases. However, BPE-patching consistently outperforms the standard BPE approach across all vocabulary sizes. Moreover, our approach demonstrates better scalability with increasing vocabulary size. This behaviour is expected, as the local encoder and decoder in our model generalize more effectively to rare words. In contrast, as the vocabulary size grows, the number of rare tokens increases, leaving many entries in the embedding matrix untrained or unused.

Furthermore, we compare the floating point operations (FLOPs) required by our model to those of the standard BPE approach. The key architectural difference lies in the use of local encoder and decoder modules in our model.
%For the BPE baseline, we consider only the FLOPs associated with the logits computation, since the embedding layer relies on indexing, which is a zero-FLOP operation.
To compute the FLOPs for a full forward pass across all layers, we follow the methodology described in Section\ref{sec:flops}.

Figure~\ref{fig:flops-increase} illustrates that our approach scales more efficiently in terms of FLOPs as the vocabulary size increases. This advantage arises from the large embedding matrix in the standard BPE approach, which incurs significant computational overhead during logits computation. In contrast, our method avoids this bottleneck by replacing the embedding matrix with local encoder-decoder modules.

\subsection{Ablation on S}
The patch size \( S \) does not affect the length of the sequence processed by the latent transformer \( f_w \); that is, global sequence compression is determined solely by the BPE tokenizer. However, \( S \) has a direct impact on the speed and memory requirements of the local transformer. In general, shorter patches are preferable—particularly during inference, which is performed autoregressively—due to reduced computational load. The tradeoff is that smaller patches limit the ability of the local transformer to model higher-level linguistic structures. Being able to provide this trade-off is part of our contrinution.
\begin{figure}
    \centering
    \includegraphics[width=0.7\linewidth]{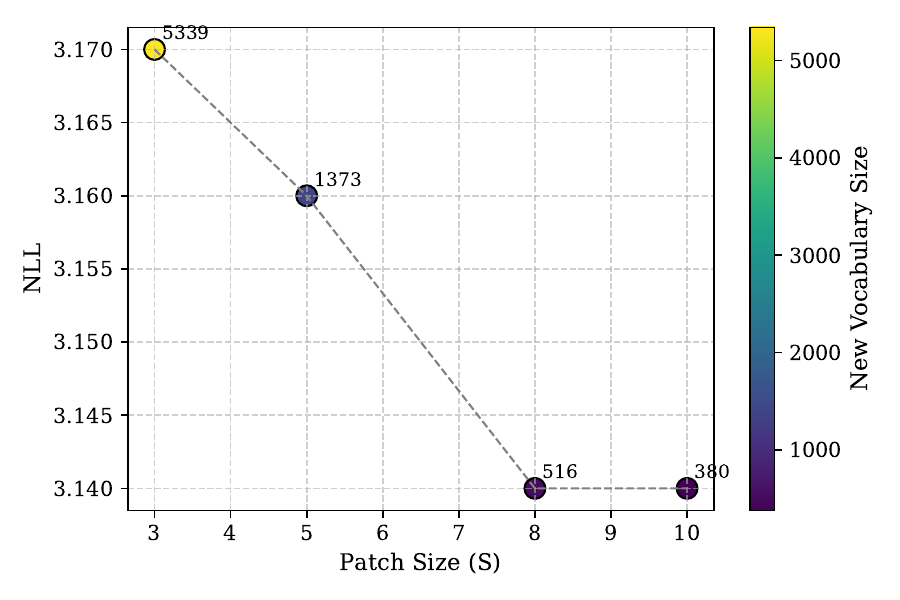}
    \caption{Ablation on the patch size (S) for a small hierarchical model.}
    \label{fig:S-ablation}
\end{figure}

We experiment with the small model, and only vary $S$ in algorithm \ref{alg:hierarchical-bpe}. We note that for patches smaller than $S$ we just use padding. The hyperparameters are the same as used in Table\ref{tab:hyperparams}. \Figref{fig:S-ablation} shows that the optimal maximum patch size in this setting is $8$. For lower values, we see a modest performance drop, while for $S=10$, the NLL stays the same. 

\subsection{Downstream Tasks}
To further verify our model we conduct experiments on question answering tasks. First, we pre-train a medium model on the next token prediction task, for 15B tokens. For the other baselines, we use the same hyperarameters as defined in Table\ref{tab:hyperparams}, alongside a maximum patch size $ S=10 $. Then we perform zero-shot evaluations on:
\begin{itemize}
    \item \textbf{Commonsense Reasoning (0-shot)}: HellaSwag~\citep{hellaswag}, PIQA~\citep{piqa}, WinoGrande~\citep{winogrande}, and ARC-e/-c~\citep{arc}.
    \item \textbf{Broad Context Understanding}: Lambada-OpenAI\citep{lambda}.
    \item \textbf{Popular Aggregated Results}: MMLU \citep{mmlu}.
\end{itemize}

To obtain the scores we simply calculate the log-likelihood per character in each patch and select the answer with the highest log-likelihood. Table~\ref{tab:bench-qa} shows that we outperform the standard BPE approach and the entropy based patching, while performing comparable to space patching in most benchmarks. 

We also evaluate the language modelling capacity of the medium model by measuring the BPB score. The results in Table~\ref{tab:bench-qa} show that our model improves with scale (better BPB score than the small model) and also outperforms the other baselines at larger sizes. 
\begin{table*}[!]   
    \centering
    \begin{tabular}{lccccccccc}
    \toprule
         Model & Param  & BPB & MMLU & Hella & Lmb. & ARC-c & ARC-e & Wino. & Piqa \\
         & & & acc  & acc\_norm & acc  & acc\_norm & acc & acc & acc \\
         \midrule
         BPE & 359M & 1.02 & 23 & 31.3 & 29.89 & \textbf{23.63} & 38.05 & 50.5 & 61.43 \\
         Ent.-Patch & 575M & 1.10 & 23.07 & 32.91 & 29.4  & 20.9 & 36.6 & 48.8 & 59.6 \\
        % MQA & \\
        Space-Patch & 575M & 1.03 & \textbf{23.24} & \textbf{35.9} & \textbf{32.8} & 21.5 & 39.02 & 52.5 & \textbf{64.2}  \\
        BPE-Patch & 323M & \textbf{0.98} & 22.9 &  34.6 & 32.5 & 22.1 & \textbf{40.6} & \textbf{53.4} & 63.3\\
        \bottomrule
    \end{tabular}
    \caption{\textbf{Common Academic Evaluation Benchmarks} comparison between standard embedding matrix and our learnable embeddings. The individual task performance is measured via zero-shot.
    }
    \label{tab:bench-qa}
    %\vskip -0.1in
\end{table*}

% \textbf{TODO: Multilingual corpus}: \url{https://arxiv.org/pdf/2404.16816}, \url{https://arxiv.org/pdf/2310.19341}

\section{Related Work}
\textbf{Model-based dynamic grouping.} Numerous prior works have explored hierarchical models, particularly those involving dynamic grouping of characters into patches. The Byte-Level Transformer (BLT)~\cite{metabyte} employs a lightweight model trained on next-character prediction to determine patch boundaries. A new patch is initiated either when the entropy is high or when the current patch exceeds a predefined maximum length. This approach eliminates the need for an explicit end-of-patch token but introduces an auxiliary model solely for patching. In contrast, our method avoids the need for such a model and handles the overhead introduced by end-of-patch tokens via a second-stage BPE compression, which effectively reduces the average patch size.

Other works, such as~\citet{DynamicTokenPool}, explore dynamic pooling through learnable models. While they also incorporate groupings informed by subword tokenizers, a key distinction is that they explicitly train a model to replicate the behavior of a BPE tokenizer—an essential step in the absence of explicit patch boundary markers. Earlier research has also attempted character-level compression using convolutional layers prior to applying a global Transformer~\cite{canine,langpixel2,langpixel1}.

\textbf{Space-based grouping.} Space patching is a well-studied method, with several recent works exploring its effectiveness at medium to large scales, including~\citet{neitemeier2025hierarchical}, \citet{spacebyte}, \citet{sun-etal-2023-characters}, and \citet{thawani2023learntokenswordpooledtokenization}. The latter introduces a minor modification to the hierarchical architecture, where the local encoder and decoder utilize more than one hidden representation per patch. Prior to space patching, grouping of consecutive characters was explored by~\citet{megabyte}. 

\textbf{Token Grouping.} Recent work has also investigated the consecutive grouping of tokens~\cite{token-grouping}. This is related to our approach in that our hierarchical BPE tokenization can be interpreted as a form of dynamic grouping, where tokens are composed from a much smaller vocabulary $V'$.
\section{Conclusion}
In this work, we presented a dynamic character grouping method that leverages BPE token boundaries to define patches, eliminating the need for an auxiliary neural network to determine patch segmentation. By augmenting the BPE process with an explicit end-of-patch token and introducing a second-stage compression step, we control patch granularity while maintaining a compact vocabulary. This design bridges the gap between character-level flexibility and subword-level efficiency, resulting in a hierarchical representation that generalizes beyond whitespace-delimited languages. Furthermore, our method can also be interpreted as a dynamic token grouping strategy. The empirical results demonstrate improved efficiency and performance over entropy-based patching and standard BPE, with competitive results relative to space-based grouping. These findings highlight the effectiveness of token-level dynamic grouping as a lightweight yet expressive alternative to conventional tokenization strategies in large language models.
\section*{Limitations}
While our method demonstrates strong performance across several benchmarks, there are key limitations worth highlighting.

First, we do not evaluate our approach in large-scale regimes involving models with billions of parameters. It remains an open question how well token-level dynamic grouping scales in such settings, where training dynamics, memory constraints, and optimization challenges may differ substantially.

Finally, our experiments are limited to the GPT-2, LLaMA3 pre-trained tokenizer and SentencePiece tokenizers trained in-house. While this setup allows us to evaluate our method in a controlled and reproducible manner, future research should investigate the impact of using other widely adopted pre-trained tokenizers—such as those used in Gemma, or PaLM—to assess the generality of our approach across different tokenizer vocabularies.

\bibliography{main}

\end{document}